# Machine Learning Approaches to Predict Breast Cancer: Bangladesh Perspective

*Taminul Islam[1], Arindom Kundu[2], Nazmul Islam Khan[3], Choyon Chandra Bonik[4], Flora Akter[5], and Md Jihadul Islam[6]

[1,2,3,4,5,6]Department of Computer Science and Engineering,
Daffodil International University,
Ashulia, Dhaka, Bangladesh
[1]taminul15-11116@diu.edu.bd

**Abstract.** Nowadays, Breast cancer has risen to become one of the most prominent causes of death in recent years. Among all malignancies, this is the most frequent and the major cause of death for women globally. Manually diagnosing this disease requires a good amount of time and expertise. Breast cancer detection is time-consuming, and the spread of the disease can be reduced by developing machine-based breast cancer predictions. In Machine learning, the system can learn from prior instances and find hard-to-detect patterns from noisy or complicated data sets using various statistical, probabilistic, and optimization approaches. This work compares several machine learning algorithms' classification accuracy, precision, sensitivity, and specificity on a newly collected dataset. In this work Decision tree, Random Forest, Logistic Regression, Naïve Bayes, and XGBoost, these five machine learning approaches have been implemented to get the best performance on our dataset. This study focuses on finding the best algorithm that can forecast breast cancer with maximum accuracy in terms of its classes. This work evaluated the quality of each algorithm's data classification in terms of efficiency and effectiveness. And also compared with other published work on this domain. After implementing the model, this study achieved the best model accuracy, 94% on Random Forest and XGBoost.

## 1    Introduction

Tumors form when a single cell divides unchecked, leading to an unwelcome growth known as cancer. Benign and malignant are the two types of classes for cancer detection. A malignant tumor develops fast and damages its tissues by invading them [1]. There is a malignant tissue that is forming in the breast that is called breast cancer. Breast cancer symptoms include an increase in breast mass, a change in breast size and form, a change in breast skin color, breast discomfort, and changes in the breast's genetic makeup. Worldwide, breast cancer is the second leading cause of death in women after heart disease. And it affects more than 8% of women at some point in their lives [2]. According to the WHO's annual report, more than 500,000 women have breast cancer every year. It is predicted that the prevalence of this disease will arise in the future due to environmental damage.

Obesity, hormone treatments therapy during menopause, a family medical history of breast cancer, a lack of physical activity, long term exposure to infrared energy, having children later in life or not at all, and early age at which first menstruation occurs are some of the risk factors for breast cancer in women. These and other factors are discussed further below. Many tests, including ultrasound, biopsy, and mammography, are performed on patients to determine whether they have breast cancer. This is because the symptoms of breast cancer vary widely. The biopsy, which includes the extraction of tissue or cell samples for analysis, is the most suggestive of these procedures.

A human observer is required to detect specific signal characteristics to monitor and diagnose illnesses. Several computer-aided-diagnosis (CAD) techniques for computer-aided diagnostic systems have been developed during the last ten years to overcome this challenge due to the enormous population of clients in the critical care section and the necessity of constant surveillance of such circumstances. Using these methods, diagnostic criteria that are predominantly qualitative are transformed into a problem of quantitative feature categorization. Diagnosis and prognosis of breast cancer results can be predicted using a variety of machine learning algorithms. This work aims to assess those algorithms' accuracy, sensitivity, specificity, and precision in terms of their efficiency and performance.

This paper has another five sections. The literature review is covered in Section 2 of this paper. Section 3 explains the methodology of this work. Section 4 covers the experimental comparison with results. Section 5 presents the findings and discussions of this work. Finally, Section 6 finishes the paper with conclusions.

## 2     Literature Review

V. Chaurasia and T. Pal applied their model to find the best machine learning algorithms to predict breast cancer. They applied SVM, Naïve Bayes, RBF NN, DT, and basic CART [3] in their work. After implementing their model that works achieved the best AUC 96.84 % on SVM in Wisconsin Breast Cancer (original) datasets.

Breast cancer survival time can be predicted using an ensemble of machine learning algorithms, as explored by Djebbari et al. [4]. Compared to earlier results, their method has a higher level of accuracy in their own breast cancer dataset. DT, SVM, Naïve Bayes, and K-NN are compared by S. Aruna and L. Nandakishore to determine the best classification in WBC [5]. They achieved the best AUC 96.99% on the SVM classifier in that work.

Tumor cells were classified using six machine learning methods developed by M. Angrap. They developed and built the Gated Recurrent Unit, a variation of the long short-term memory neural network (GRU). neural network softmax layer was replaced with the support-vector machine layer (SVM). GRU SVM's 99.04 % accuracy was the best on that work [6]. Utilizing association rules and a neural network to train the model, Karabatak et al. [7] increased the model's accuracy to 95.6% by using cross-validation. It was used Naïve Bayes classifiers with a new technique for weight modification.

Mohebian et al. [8] investigated the use of ensemble learning to predict the recurrence of cancer. Researchers Gayathri et al. conducted an evaluation of three machine learning models that had the most significant outcomes when utilizing a relevance vector [9]. To get the best results, Payam et al. used preprocessing and data reduction techniques, such as the radial basis function network (RBFN), in combination [10].

Breast cancer survival prediction models were developed using data from research on breast cancer published in [11]. Breast cancer survivorship prediction algorithms were used for both benign and malignant tumors in this work. ML algorithms for breast cancer detection have been studied extensively in the past, as shown in [12]. They proposed that data augmentation approaches might help alleviate the issue of a small amount of data being available. Using computer-aided mammography image characteristics, the authors in [13] demonstrated a method for detecting and identifying cell structure in automated systems. According to [14], numerous classification and clustering techniques have been compared in the study. Classification algorithms outperform clustering algorithms, according to the findings. Table 1 states the clear comparison between this work and other previously published work.

**Table 1.** Comparison with previously published work

| Ref | Year | Contribution | Dataset | Algorithms | Best Accuracy |
|-----|------|--------------|---------|------------|---------------|
| [15] | 2020 | Implemented machine learning techniques to predict breast cancer. | UCI machine learning repository | ANN, DT, SVM, and NB. | 86% |
| [16] | 2020 | Developed a model to predict breast cancer. | UCI machine learning repository | RF, XGBoost | 74.73% |

| [17] | 2020 | Predicted breast cancer using effective data mining classifiers. | Wisconsin Breast Cancer dataset | K-Means Clustering, DT | 94.46% |
|------|------|---|---|---|---|
| [18] | 2018 | Implemented machine learning techniques to predict breast cancer. | Wisconsin Breast Cancer dataset | KNN, NB, SVM, RF | 97.9% |
| [19] | 2019 | Predicted whether the person has breast cancer or not. | Multidimensional heterogeneous data. | KNN, SVM, RF, GB | 93% |
| [20] | 2021 | Predict breast cancer at an early stage or malignant stage. | Wisconsin Diagnostic dataset | SVM,K-NN, NB, DT, K-Means, ANN | 97% |
| [21] | 2017 | The principle component analysis (PCA) approach is applied to successfully increase the moral rectitude of the attributes addressing eigenvector issue. | Wisconsin Diagnostic dataset. | SVM | 93% |
| [22] | 2018 | Implemented model to classify tumor as benign or malignant. | Wisconsin Diagnostic dataset. | RF, K-NN, NB | 95% |

## 3   Methodology

An overview of the main technique of the study is provided in this section. The main workflow of this research is shown in Fig. 1. Dataset origin and features are discussed in this section. In addition, the surrounding background is discussed. This chapter concludes with a brief discussion of specific classification models and assessment techniques. Our researchers have collected the data manually. Then It is important to clean up noisy and inconsistent data using preprocessing techniques. Various preprocessing techniques have boosted the final performance of this work. We also had to eliminate erroneous data from the model to get it to work. For trials relevant to this study, we train numerous five classification algorithms. Fig. 1. illustrates the main workflow of this research.

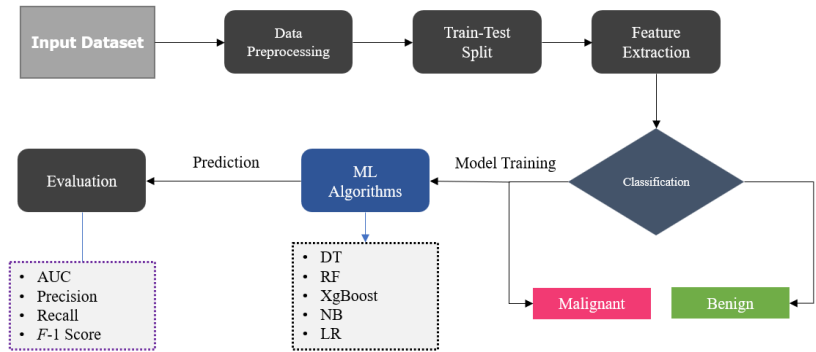

**Fig. 1.** Proposed model workflow

## 3.1 Data Description

In this work, researchers have collected total of 456 data from three hospitals of Bangladesh named Dhaka Medical College Hospital, LABAID Specialized Hospital, and Anwar Khan Medical College Hospital. The data contains 254 benign and rest 202 are classified as malignant that shown in Fig. 2.

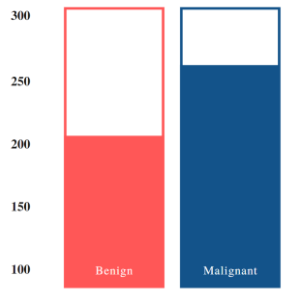

**Fig. 2.** Data class distribution

The entire dataset was factored in when doing the dataset analysis. It is counter-plotted in Fig. 3. that the dataset's mean radius feature. Patients suspected of having cancer have a radius greater than 1, whereas those who don't appear to have the disease have a radius closer to 1.

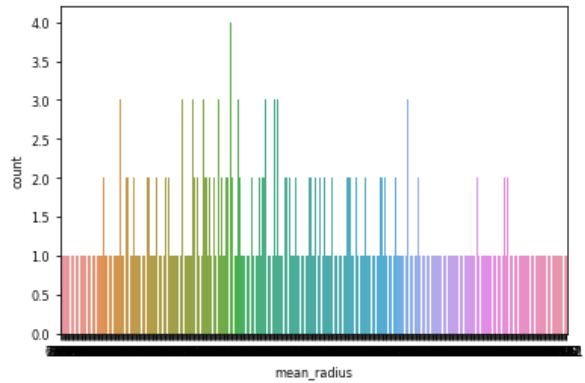

**Fig. 3.** Mean radius of the dataset.

Heatmaps are shown in Fig. 4. to highlight the association between the dataset's characteristics. 2D Correlation Matrix Correlation Heatmap illustrates a two-dimensional matrix between two discrete dimensions, where each row and each column represents one of the two-dimension values. Colored pixels on a monochrome scale are used in this heatmap to highlight the correlation between the dataset's attributes. There is a growing correlation as the intensity of the color increases. The number of observations that meet the dimensional values is directly proportional to the color value of the cells. The proportionality between the two characteristics is used to determine the dimensional value. The positive correlation is obtained when both variables vary and move in the direction. A reduction in one measure is correlated with a rise in the other, and the opposite is true. There are six features in this work. These are: mean radius, mean perimeter, mean texture, mean smoothness, mean area, and diagnosis.

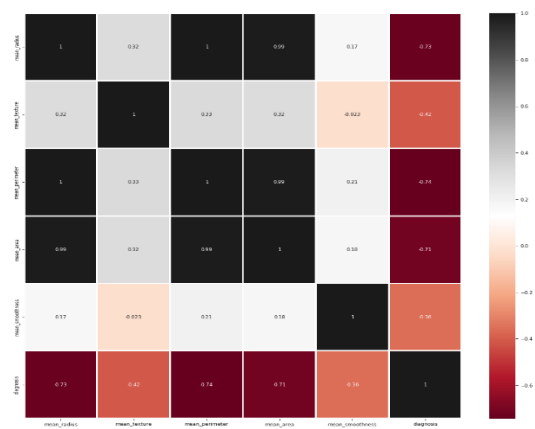

**Fig. 4.** Correlation between the features.

### 3.2 Data Preprocessing

To begin a study, it is necessary to preprocess the data that has been collected. Our first step is to analyze the information we've gathered so far. For this reason, we rely on a variety of sources for our data. To begin, we prepare the data needed to remedy the issue. These data sets contain a wide variety of numerical values. A single piece of this data is analyzed at a time. Machine learning models can process only numerical data. Before data analysis, the mean and mode were trimmed. Before computing the average, the highest and most minor numbers are trimmed by a tiny percentage [23]. The percentage measurements are trimmed in two directions.

### 3.3 Machine Learning Models

Machine learning is the most accessible way to predict breast cancer disease. In the Literature Review section, it is clear that the maximum of the work has been done successfully by machine learning and deep learning methods. We know deep learning is a subset of machine learning. Five different machine learning algorithms have processed this new dataset to discover the most accurate method. Decision Tree, Naïve Bayes, Extreme Gradient Boosting, Logistic Regression, and Random Forest are all classifiers of these techniques. These models have gained a brief emphasis in this section.

#### 3.3.1 Decision Tree (DT)

DT is a robust machine learning algorithm that can classify and predict data [24]. For the vector to go ahead, each node acts as a test criterion, and the terminal nodes offer a projected class or prediction value. DT can be structured in this way: DT works well for a small number of class labels, but it doesn't work as well when there are many classes and a low number of training data. Additionally, the computational cost of training DTs can be significant.

#### 3.3.2 Random Forest (RF)

Many different paths can be explored in RF. The amount of trees in the forest has a direct bearing on the outcome. The more trees we have, the more accurate our results will be. The classifier in RF is either C4.5 or J48. Bagging and various feature selection for decision trees were proposed by Breiman in 2001 [25]. RF is a classifier that requires human supervision.

#### 3.3.3 Extreme Gradient Boosting (XGBoost)

Gradient-boosting frameworks employ a method called XGBoost, which is an ensemble decision-tree technique. In general, decision trees are simple to see and understand, but developing an intuitive understanding for another era of tree-based algorithms can be challenging [26].

### 3.3.4 Naïve Bayes (NB)

Naïve Bayes is a classifier that considers that each feature only influences the class [15]. As a result, each feature is merely a child of the class. NB is appealing because it provides a theoretical foundation that is both explicit and solid, ensuring the best possible induction from a given set of assumptions. According to several real-world examples, the total independence assumptions of the features concerning the class are broken. Inside the wake of such violations, however, NB has shown to be extraordinarily strong. Thanks to its straightforward structure, NB is quick, easy to deploy, and successful. Useful for high-dimensional data, as each feature's probability is evaluated separately [16].

### 3.3.5 Logistic Regression (LR)

Logistic regression is a supervised learning classification approach. X can only have discrete impacts on the classification problem's target variable (or output), y. Logistic regression is, in fact, a regression model. It constructs a regression method to forecast the probability data input falls into the "1" category [27]. Using logistic regression, classification challenges such as cancer detection can be addressed quickly.

### 3.4 Experimental Setup

A training and testing phase were used to implement all five machine learning algorithms. The dataset was separated with distinct values assigned to the algorithm's data selection to train and test the model. We sent 80% data for training and 20% for testing. The experiment was run on a Jupyter notebook running Python 3.0 with 12GB RAM on an Intel Core i5 10th generation CPU. The experiment was conducted using the sklearn library such as pandas, TensorFlow, matplotlib, and Keras.

### 3.5 Performance Measurement Unit

Training and generalization errors are two of the most common types of mistakes. Increasing the complexity of the model can help reduce training errors since the complexity of the model reduces the training error rate. To minimize generalization errors, use the Bias–Variance Decomposition (Bias+Variance) technique. It's called overfitting if a drop-in training error leads to a rise in test error rates. Accuracy, precision, recall, and $F_1$-Score can be used to evaluate the performance of each categorization system.

Various authors employed a diverse variety of to evaluate their models' efficacy.

Even though the bulk of the research utilized many indicators to evaluate their performance, a minor number of studies also used a single statistic to do the same. In this work, accuracy, precision, recall & $F_1$-Score is considered for evaluating this research work. This four-measurement unit is the best for prediction data analysis.

Accuracy relates to the capability to identify and classify instances [18] correctly.

Here TP = True Positive, TN = True Negative, FP = False

Positive and FN = False Negative. Equation 1 shows the mathematical expression of accuracy.

$$\text{Accuracy} = \frac{\text{TP+TN}}{\text{TP+FP+TN+FN}} \quad (1)$$

For statistical analyses, precision is defined as the number of observed positive events divided by the total number of expected positive events [17]. Equation 2 shows the mathematical expression of precision.

$$\text{Precision} = \frac{\text{TP}}{\text{TP+FP}} \quad (2)$$

The model's recall measures how well it can identify those people who have cancer [17]. Equation 3 shows the mathematical expression of recall.

$$\text{Recall} = \frac{\text{TP}}{\text{TP+ FN}} \quad (3)$$

Due to its reliance on both precision and recall, this is referred to as the harmonic mean. Equation 4 is an expression of a mathematical equation for $F_1$ Score [17].

$$F_1 \text{ Score} = 2 \left( \frac{\text{Precision} \times \text{Recall}}{\text{Precision+Recall}} \right) \quad (4)$$

## 4    Experimental Evaluation

There are a total of five machine learning methods that have been applied to this dataset. When comparing the performance of one algorithm to another, there is a tight difference found. Based on the accuracy level RF and XGBoost performs better than the other five algorithms. RF and XGBoost achieved the best 94% accuracy, whereas NB and LR achieved equally 93% accuracy. This study found 91% accuracy in DT. In this study, we found the best precision from the NB and LR, but in terms of overall accuracy, NB and LR stand jointly third position. On the other hand, XGBoost and RF provide the best recall 0.98 and 0.97 among the other algorithms. This work found the lowest performance from the DT so far. Table 2 illustrates the classification report comparison of five machine learning algorithms, where each method is evaluated in terms of its performance in two classes: benign and malignant.

**Table 2.** Comparison of classification report among five algorithms.

| Algorithms | Class | Precision | Recall | $F_1$ Score | Accuracy |
|---|---|---|---|---|---|
| DT | Benign | 0.91 | 0.88 | 0.89 | 0.91 |
| | Malignant | 0.91 | 0.94 | 0.93 | |
| RF | Benign | 0.96 | 0.90 | 0.92 | 0.94 |
| | Malignant | 0.93 | 0.97 | 0.95 | |
| NB | Benign | 0.83 | 1.00 | 0.91 | 0.93 |
| | Malignant | 1.00 | 0.89 | 0.94 | |
| XGBoost | Benign | 0.98 | 0.88 | 0.92 | 0.94 |
| | Malignant | 0.92 | 0.98 | 0.95 | |
| LR | Benign | 1.00 | 0.83 | 0.91 | 0.93 |
| | Malignant | 0.89 | 1.00 | 0.94 | |

In the Fig. 4. it it clearly shows that, the accuracy of RF and XGBoost is higher than other five machine learning methods, as demonstrated by the graph. Fig. 4. shows the accuracy percentage between applied five machine learning algorithms. Here it is clearly shows that, DT performs poor than all applied algorithms where NB and LR performs equally same. Finally RF and XGBoost perform best in terms of AUC comparisn.

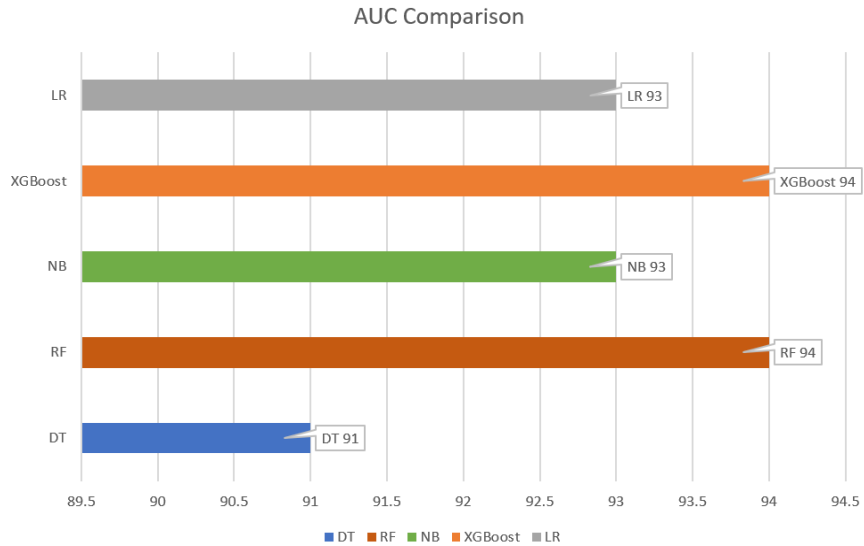

**Fig. 4.** AUC comparison of five machine learning algorithm.

A classification algorithm's performance can be summarized easily using a confusion matrix. Even if your dataset has just two classes, the accuracy of your classification might be deceiving if the number of observations for each class is uneven. We can better understand how accurate the classification model is by calculating a confusion matrix (CM) [19].

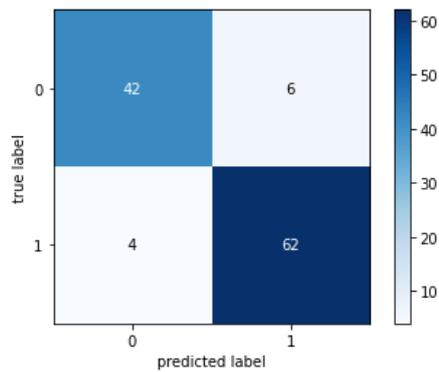

**Fig. 5.** Confusion matrix of DT

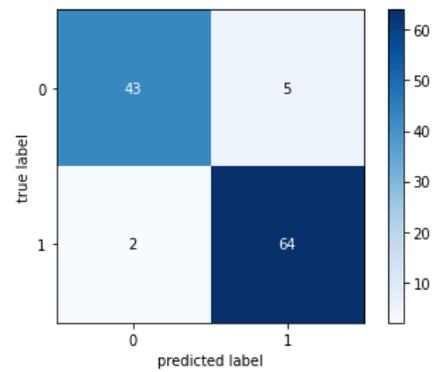

**Fig. 6.** Confusion matrix of RF

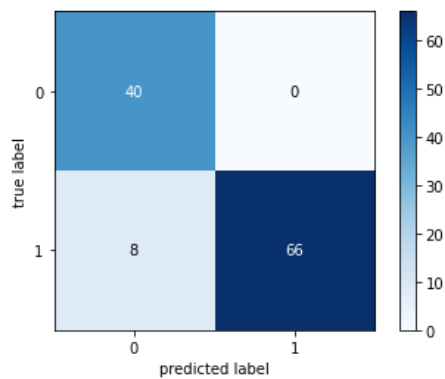

**Fig. 7.** Confusion matrix of NB

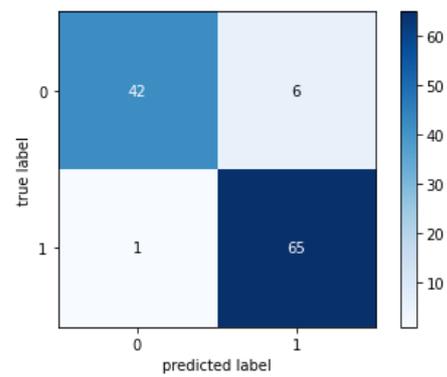

**Fig. 8.** Confusion matrix of XGBoost

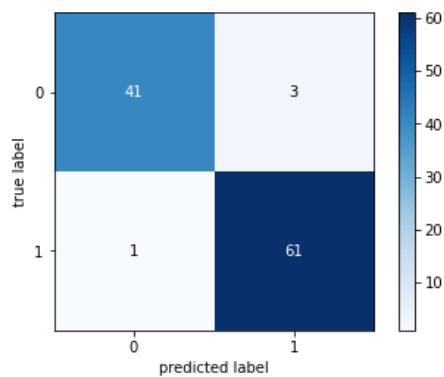

**Fig. 9.** Confusion matrix of LR

Fig. 5. – Fig. 9. Show the five confusion matrix to present the performances of applied five machine learning algorithms where x-axis states the predicted level and y-axis states the true level.

## 5 Discussion

The majority of the current research in this domain is devoted to improving the ability to predict the incidence of breast cancer. But few works have been done with a newly collected dataset. Machine learning algorithms were used to see if they could better and predict cancer with the best possible accuracy. The classification strategy worked effectively in this work. It is essential to compare this work with other works to present its contribution to the global society.

The main goal of this work was to find the best machine learning techniques that can predict breast cancer with maximum performance. Breast cancer prediction is very alarming work. That's why there is a lot of work on this domain. But maximum of the work has been done with a particular two or three open accessed datasets. For this reason, we found an almost similar difference between one to another algorithms. Besides, the AUC score was quite similar between various published works. UCI machine learning repository and Wisconsin Diagnostic dataset are the common datasets used on a maximum of the work. Those work found almost above 90% of accuracy. There are different numbers of AUC on breast cancer prediction where this research work was compared. This work found 94% best accuracy on Rf and XGBoost algorithm. After pre-processing all the datas, we did the feature extraction. Then we applied the algorithms. This work has applied five machine learning algorithms with a newly collected dataset. There is the major difference between other published work. However, In terms of a new collected dataset we found the best accuracy on Random Forest and XGBoost algorithm. The accuracy level can be high in future with more accurate and balanced dataset.

## 6 Conclusion

In Bangladesh, breast cancer is the most dangerous disease for women that stands at the top level for its death ratio. There are several machine learning and data mining techniques that use to examine medical analysis. Classifiers for medical diagnostics that are both accurate and efficient in computing provide a significant challenge for data miners and machine learning researchers. This work has applied five leading machine learning algorithms, DT, RF, XGBoost, NB, and LR, to predict breast cancer on a new dataset. The dataset has been collected from three Bangladeshi hospitals. After implementing the model, this work achieved the best 94% accuracy on RF and XGBoost algorithms. We compared the findings with those of previous research and discovered work, and we found that this approach performed well. This research relied heavily on datasets and methodology. There are some limitations on this work. Collecting all quality new data in this pandemic circumstance was difficult. There can be more data to collect. But in terms of the new dataset, the outcome satisfies us. We should employ more efficient training and preprocessing approaches to achieve better results. Increasing the sample size of the dataset in the future will ensure the accuracy and effectiveness of this study.